\documentclass{article}

\pdfoutput=1

\usepackage[preprint, nonatbib]{neurips_2021}
\usepackage[numbers]{natbib}
\usepackage[utf8]{inputenc} 
\usepackage[T1]{fontenc}    
\usepackage{hyperref}       
\usepackage{url}            
\usepackage{booktabs}       
\usepackage{amsfonts}       
\usepackage{nicefrac}       
\usepackage{microtype}      
\usepackage{xcolor}         
\usepackage{amsthm,amsmath}
\usepackage{amssymb}
\usepackage{mathtools}
\usepackage{algorithm,algorithmic}
\usepackage{graphicx} 
\usepackage{enumerate}
\usepackage{enumitem}
\usepackage{wrapfig}
\usepackage{amsmath}
\usepackage{amssymb}
\usepackage{graphicx}
\usepackage{bbm}
\usepackage{booktabs}

\newcommand{\vx}{{\mathbf{x}}}
\newcommand{\vy}{{\mathbf{y}}}
\newcommand{\vz}{{\mathbf{z}}}

\newcommand{\cA}{{\mathcal{A}}}
\newcommand{\cB}{{\mathcal{B}}}

\newcommand{\cD}{{\mathcal{D}}}

\newcommand{\cL}{{\mathcal{L}}}

\newcommand{\cN}{{\mathcal{N}}}

\newcommand{\cP}{{\mathcal{P}}}

\newcommand{\cS}{{\mathcal{S}}}
\newcommand{\cT}{{\mathcal{T}}}

\newcommand{\RR}{\mathbb{R}}

\newcommand{\EE}{\mathbb{E}}

\DeclareMathOperator*{\argmin}{arg\,min}

\newcommand{\bc}{\begin{center}}
\newcommand{\ec}{\end{center}}
\newcommand{\bdm}{\begin{displaymath}}
\newcommand{\edm}{\end{displaymath}}
\newcommand{\beq}{\begin{equation}}
\newcommand{\eeq}{\end{equation}}
\newcommand{\bfl}{\begin{flushleft}}
\newcommand{\efl}{\end{flushleft}}
\newcommand{\bt}{\begin{tabbing}}
\newcommand{\et}{\end{tabbing}}
\newcommand{\beqn}{\begin{align}}
\newcommand{\eeqn}{\end{align}}
\newcommand{\beqs}{\begin{align*}} 
\newcommand{\eeqs}{\end{align*}}  

\newtheorem{mythm}{Theorem}

\newtheorem{problem}{Problem}

\usepackage[utf8]{inputenc} 
\usepackage[T1]{fontenc}    
\usepackage{url}            
\usepackage{booktabs}       
\usepackage{amsfonts}       
\usepackage{nicefrac}       
\usepackage{microtype}      
\usepackage{subcaption}
\usepackage{color}
\begin{document}
\raggedbottom
\setlength{\belowcaptionskip}{-10pt}

\title{Persona Authentication through Generative Dialogue}
\author{%
  Fengyi Tang \\
  Michigan State University\\
  \texttt{tangfeng@msu.edu} \\
  \And
  Lifan Zeng \\
  Nuro.ai \\
  \texttt{zenglifa@msu.edu} \\
  \AND
  Fei Wang\\
  Weill Cornell Medicine \\
  \texttt{feiwang03@gmail.com} \\
  \And
  Jiayu Zhou \\
  Michigan State University\\
  \texttt{tangfeng@msu.edu} \\
}

\maketitle

\begin{abstract}
In this paper 
we define and investigate the problem of \emph{persona authentication}: learning a conversational policy to verify the consistency of persona models.  
We propose a learning objective and prove (under some mild assumptions) that local density estimators trained under this objective maximize the mutual information between persona information and dialog trajectory. 
Based on the proposed objective, we develop a method of learning an authentication model that adaptively outputs personalized questions to reveal the underlying persona of its partner throughout the course of multi-turn conversation. 
Experiments show that our authentication method discovers effective question sequences that generalize to unseen persona profiles. 
\end{abstract}

\section{Introduction}
\label{sec:intro}
In recent years, 
one promising approach to diverse and personalized dialog generation has been  \textit{persona} models~\cite{li2016persona,zhang2018personalizing,song2019exploiting} which embed the so-called ``persona'' information (e.g., name, gender, and self-descriptions) into neural conversational agents.
While the goal of persona modeling is to achieve human-level response diversity and character consistency, 
a critical yet often overlooked factor 
is the \emph{sequence of prompts} used to induce diversity in generated responses.

Consider the toy example in Table~\ref{table:persona_example}. 
Certain sequences of question prompts (from the authenticator) 
create a trail of generated responses that reveal more persona information, compared to non-specific conversation (random policy). 
Yet it is unclear \emph{a priori} which sequence(s) of questions most effectively reveal the dialog agent's underlying persona. 
Moreover, a set of questions may be effective for one persona but fail for others. 
Currently, long interactions with humans are necessary to gain insight into persona model characteristics such as authenticity~\cite{song2019exploiting,yu2016strategy}, diversity~\cite{zhang2018personalizing}, and engagement~\cite{song2019exploiting}. 

In this paper we present a learning approach for interacting with conversational agents. 
Specifically, we introduce the \textbf{persona authentication} problem, where a model estimates the persona information of an input agent by learning to deliver a sequence of questions that progressively reveal more information about the agent throughout the course of the dialog.
This is difficult because 
exact search through the space of possible question sequences is infeasible. 
Therefore a model must adaptively prune its set of potential questions based on the dialogue agent's responses. 

\begin{table*}[ht]
\footnotesize
\centering
\begin{tabular}{c c c}\\\hline
\multicolumn{3}{c}{\textbf{Persona Descriptions}}\\  \hline
1. I am a construction worker. & 2. I enjoy building houses. &
3. I have 5 cats that are very special to me. \\ \hline
\end{tabular}
\begin{tabular}{c c | c c} \\ \hline
\textbf{Role} & \textbf{Response} & \textbf{Role} & \textbf{Response} \\ \hline
Authenticator &  \multicolumn{1}{p{4cm}|}{hello! what kind of work do you do?} & Random Policy & hello how are you today? \\
Persona Model & i build houses. & Persona Model &  \multicolumn{1}{p{4cm}}{great! i just got back from work.}\\\hline
Authenticator & \multicolumn{1}{p{4cm}|}{that's awesome. what do you do outside of work?} & Random Policy & \multicolumn{1}{p{4cm}}{me too. i'm a teacher at a high school.}\\
Persona Model & i like to spend time with my cats. & Persona Model & cool, what grade do you teach? \\\hline
\end{tabular}
\caption{Persona model responses can differ greatly depending on input questions.}
\label{table:persona_example}
\vspace{-0.1in}
\end{table*}

We further decompose persona authentication into two parts: 
\emph{persona identification}, which is inferring a set of persona features from a given dialog trajectory, and \emph{persona verification}, the problem of finding a second conversational model -- we call it a \emph{question policy} -- to elicit dialog trajectories for persona identification. 
To address the intractability of exact search through the space of dialog trajectories, we introduce a computationally tractable algorithm and show its asymptotic convergence (in cumulative conversations) toward the full persona identification objective. 
The key contributions of this paper can be summarized as follows:
\begin{itemize}[leftmargin=.1in]
    \item We introduce the \emph{authentication} loss and show that estimators trained to convergence under this objective maximize the mutual information between dialog history and persona.
    \item Based on the authentication loss we learn a dialog verification model that effectively generates question sequences to distinguish the persona of input models. 
    Empirically, we show that the question policies of the verification model adapt to out-of-distribution personas.
    \item We present a way to incorporate question policies into language model (LM) based dialog models, e.g., GPT-2, without sacrificing the felicity and consistency of the original LM model.
\end{itemize}


\section{Persona Authentication}
\label{sec:problem}
%
\subsection{Notation}
Let $\cD = \{ \tau_i \}_{i=1}^n$ be a set of dyadic dialog samples. 
Each dialog follows the form $\tau = \{X, Y \}$, where $X = (X_t)_{t=1}^T$ denotes the sequence of \emph{source responses} and $Y = (Y_t)_{t=1}^T$ denotes the sequence of \emph{target responses}. 
Each response is composed of a sequence of tokens, represented as $(\vx_k^{(t)})_{k=1}^K$ (source tokens) and $(\vy_k^{(t)})_{k=1}^K$ (target tokens).
To be consistent with state-of-the-art (SOTA) dialog model  decoders~\cite{zhang2019dialogpt,wolf-etal-2020-transformers}, we use Byte-Pair Encoding (BPE)~\cite{shibata1999byte} for tokenization. 
Additionally, $T$ signifies the maximum number of turns in a dialogue sample, $K$ the maximum number of tokens per response. 
%
As a shorthand, we write $\tau_{t}$ to denote the \emph{dialogue trajectory} $(X_1, Y_1, \dots, X_t, Y_t)$ up to turn $t$, with $Y_{1:t}$ to signify the sequence of responses $Y_1, \dots, Y_t$.
Similarly $\vy_{1:k}$ represents the ordered sequence of tokens up to token $\vy_k$.

\subsection{Persona Identification} 
%
The standard objective of persona models is:
{
\begin{align}\max_{Y_t}~\log p(Y_t | X_t, \tau_{1:t-1}, P_Y ),
\label{eqn:persona_transition} 
\end{align}}
where $P_Y$ is the set of persona descriptions for the dialog agent.
Zhang et al.~\cite{zhang2018personalizing} and the \texttt{ConvAI2} challenges~\cite{dinan2020second} provided numerous ways to incorporate persona information into the dialogue generation process. 
Recently, generative persona models~\cite{wolf2019transfertransfo,zhang2019dialogpt} 
%
have been shown to be effective at  contextualized decoding by incorporating persona $P$ as language model context. 
Due to their effectiveness, we only consider generative persona models in this paper. 

To identify a persona from a given trajectory,
we formulate the \emph{persona identification} problem: 
\begin{problem}{Persona Identification.}
Given an input dialogue trajectory $\tau$, find the persona $P$ that maximizes the mutual information between $P$ and $\tau$.
More formally, the optimization objective is 
{
\begin{align}
\max_{P}~I(P, \tau) = \max_P~H(\tau) - H(\tau | P) =  \min_{P}~H(\tau |P),
\label{eqn:identification}
\end{align}}
where $H(\cdot)$ is entropy and $P \in \RR^m$ is a vector in the space of possible personas. 
\label{thm:identification}
\end{problem}
Persona identification seeks a fixed-length representation of persona information that captures the consistency of generated responses. In other words, a personalized dialog agent has to not only generate \emph{diverse} responses (high entropy $H(\tau)$), but it must also stay \emph{consistent} to a persona profile throughout multiple turns of conversation, minimizing $H(\tau|P)$. 
One challenge is that 
it is unclear how to arrive at a set of questions $X_{1:T}$ to generate the input trajectory $\tau$.
For example, certain sets of questions may always result in generic responses, regardless of the agent quality. 
Thus, problem~\ref{thm:identification} requires a way to constrain the \emph{question policy} so that, given the right set of questions, the persona of the dialog agent can be elicited. 

\subsection{Persona Verification}
We addresses the above issue by formulating question generation as an optimization problem. 
The inputs are dialog agents, i.e. trained persona models parameterized by different personas. 
We define the \emph{persona verification} problem as follows:
\begin{problem}{Persona Verification.} Given a space of persona information $\cP$, 
 persona verification is the optimization objective:
 {
 \begin{align}
    \min_{\theta}~\EE_{P \sim \cP}\big[ \cL(\tau_\theta, P) \big]
    \label{eqn:verification}
 \end{align}}
where $\cL( \cdot, \cdot)$ is the authentication loss: 
{
\begin{align}
\cL(\tau_\theta, P) = \max\{ 0, C + d(\tau_\theta, P^+) & - d(\tau_\theta, P^-) \}  - \log p(\tau_\theta).
\label{eqn:authentication}
\end{align}}
$P^+$ denotes persona facts that co-occur with trajectory $\tau$, $P^-$ the opposite. 
$C$ specifies the desired margin of separation, $d(\cdot, \cdot)$ outputs an embedding distance and 
$\tau_\theta$ is the dialog trajectory generated by the question policy ($\theta$).
\label{thm:authentication}
\end{problem} 
The first term of Eqn.~(\ref{eqn:authentication}) approximates  Eqn.~(\ref{eqn:identification}) through a triplet loss using negative sampling over the space of possible personas.
The rationale behind the first term is to address the intractability of solving for Eqn.~(\ref{eqn:identification}) directly. 
In Section~\ref{sec:analysis}, we show that this triplet loss component converges to the mutual information term in Eqn.~(\ref{eqn:identification}).
%
The second term in Eqn.~(\ref{eqn:authentication}) gives the likelihood of the trajectory. 
In order to minimize the second term, a verification algorithm has to generate queries with high likelihood under a given language model, e.g. GPT-2. 
If either the input agent or the question policy generate nonsensical responses, then the resulting $p(\tau_\theta)$ will be close to zero.
For the rest of the paper, we will refer to ``identifier'' as a model used to solve the identification problem and ``verifier'' to denote a model used to solve the verification problem.

\subsection{Analysis of Persona Authentication Objective}
\label{sec:analysis}
Now we analyze the relationship between Eqn.~(\ref{eqn:authentication}) and the mutual information between $P$ and $\tau$. 
First, we assume that for a given persona $P$, the density function for $p(\tau | P)$ follows the probability density function (PDF) of a Gibbs distribution:
{
\begin{align*}p(\tau | P) = \frac{\exp[-\beta E(\tau, P)]}{ \int_{\tau' \in \cD} \exp[-\beta E(\tau' , P)]},\end{align*}}
where $E(\tau ,P)$ is an energy function which scores the un-normalized co-occurrence likelihood of a specific dialogue trajectory $\tau$ and persona $P$. 
$\beta$ is the temperature term which controls the overall entropy of the distribution. 
We choose the Gibbs distribution because of its expressiveness and common use in contrastive learning~\cite{lecun2006tutorial}. 
Then we can express the mutual information between $\tau$ and $P$ as:
{
\begin{align}
    I(\tau, P) & = H(\tau) -  H(\tau | P) = -\EE_\tau[\log p(\tau)] + \EE_{\tau,P}\bigg[ \beta E(\tau, P) -  \log \int_{\tau' \in \cD} \exp[-\beta E(\tau', P)] \bigg] .
    \label{eqn:mutual_info}
\end{align}}
\noindent In Eqn.~(\ref{eqn:mutual_info}), the first term on the RHS corresponds to the entropy of dialog trajectories (diversity of generated responses), which is determined by the decoding quality of the input agent. 
The second term depends on our question policy and our estimation of $P$.  Trajectories under $P$ depend on the question policy since the input agent maximizes $p(Y_t | X_t, \tau_{t-1}, P)$. 
Since $P$ is not known by the policy beforehand, it is estimated each turn by the identifier. 
Unfortunately, directly estimating the entire second term is difficult -- the partition function of the conditional distribution requires us to integrate over the space of trajectories, an intractable task.

We thus propose a local density estimation of the conditional density $p(\tau | P)$ as follows: 
let
{
\begin{align}
    \widetilde{P}_\cN = \frac{1}{n V(C_n)} \sum\nolimits_{i=1}^n K(\varphi(\tau_i), \psi(P), C_n),
    \label{eqn:local_density}
\end{align}}
denote an empirical estimate of $p(\tau|P)$ using $n$ sampled trajectories. $\psi$ and $\varphi$ are embedding representations of $P$ and $\tau$, respectively. $V(C_n) = \int_{\cN_{C_n}(P)} dP'$ gives the volume of a neighborhood ball of radius $C_n$ around $\psi(P)$. $K$ is a kernel function (which we show to be a valid kernel function in Supplemental Materials) designed as follows: 
\begin{align}
    K(\tau, P, C_n) &= \begin{cases} 
    1 &\text{if } d(\varphi(\tau), \psi(P)) \leq C_n  \\ 0 & \text{else}  \end{cases}.
    \label{eqn:kernel_form}
\end{align}
%
We now present the main theorem of our analysis. 
%
\begin{mythm}{(Convergence of $\widetilde{P}_N$)}\\
If Eqn.~\ref{eqn:authentication} (authentication loss) is minimized with $0$ loss over $\cD = \{\tau_i\}_{i=1}^n$ and $\cP = \{P_j\}_{j=1}^m$, then $\widetilde{P}_N$ asymptotically converges to $p(\tau | P)$, i.e., 
\begin{align}
\lim_{n\rightarrow \infty}~\widetilde{P}_N = p(\tau | P)
\label{eqn:convergence}
\end{align}
when the following conditions hold:
\begin{align}
    \lim_{n\rightarrow \infty}~n V(C_n) = \infty, \quad \lim_{n\rightarrow \infty}~V(C_n) = 0, \quad
    \lim_{n\rightarrow \infty}~\frac{k}{n}  = 0,
    \label{eqn:convergence_conditions}
\end{align}
where $k$ is the expected number of samples that fall within $N_{C_n}(P).$
\label{thm:convergence}
\end{mythm}
The proof is provided in the Supplemental Materials. 
The goal of the identifier model is to learn the embedding functions $\varphi$ and $\psi$: 
{
\begin{align}\varphi, \psi = \argmin_{\varphi, \psi}~\frac{1}{nk} \sum_{i=1}^n \sum_{j=1}^k \max~\{0,  C + d(\varphi(\tau), \psi(P_i)) - d(\varphi(\tau), \psi(P_j)) \},
\label{eqn:analysis_contrastive_loss}
\end{align}}
where $P_i \in \cP^+$, and each $P_j$ belongs to the set of $k$ negative persona samples. 

\section{Methodology}
\label{sec:method}
\begin{figure*}
    \centering
    \includegraphics[width=.95\textwidth]{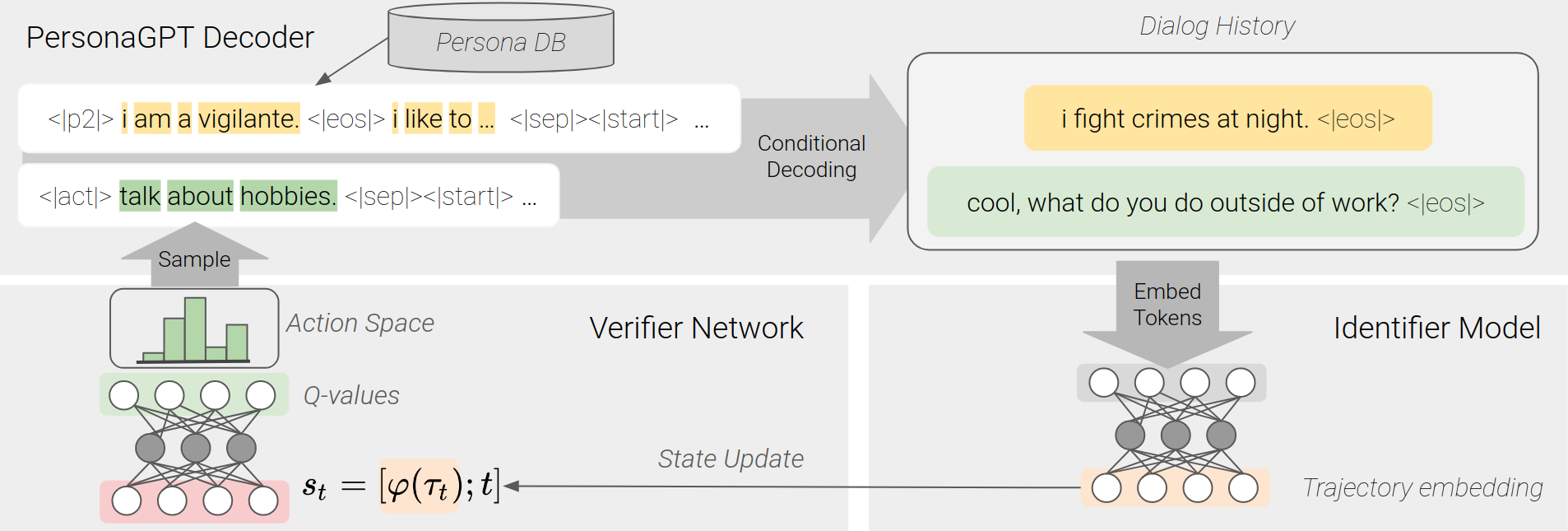}
    \caption{Overview of authentication pipeline. The identifier model represents the current conversation history as state input to the verifier. The verifier outputs a distribution $\pi(\cdot|s_t)$ over the action space. The sampled action from $\pi(\cdot|s_t)$ is converted to a question code and incorporated into PersonaGPT to decode the next question. }
    \label{fig:overview}
\end{figure*}
Figure~\ref{fig:overview} summarizes the key components of our authentication pipeline. 
We will refer to the full authentication pipeline as ``authenticator'' for short.
Once we can estimate $P$ based on the learned representation $\varphi$ of the dialog history, we can learn a question policy $\theta$ under the authentication objective (Eqn.~\ref{eqn:authentication}). 
Toward that goal, we first present an effective way to do incorporate the question policy i.e. $(X_t)_{t=1}^T$, as context for conditional decoding using SOTA LM-based dialog models, e.g. the GPT-2 architecture. 
This requires us to define an action space of control codes~\cite{keskar2019ctrl} to contextualize the decoder during question generation. 
We describe an active learning approach to learn such an action space.  
Then, we present the verifier model details and provide an algorithm for learning the question policy. 
\subsection{Action Space as Control Codes}
Since the goal of the verifier is to generate a sequence of questions $(X_t)_{t=1}^T$, we can describe the \emph{token-level} likelihood of each question $X_t$ as:
{
\begin{align}
    p(\vx_{1:k}^{(t)} | X_t, & \tau_{t-1}, P)  \quad \approx \quad p(\underbrace{ \vx_{1:k}^{(t)}}_{\text{decoded tokens}}| \underbrace{X_t}_{\text{question at }t}, \underbrace{\varphi(\tau_{t-1})}_{\text{history embedding}})
    \label{eqn:token_annot}
\end{align}
}
Unlike the persona model, the verifier does not have access to the actual persona $P$ of the input model. 
Instead, the identifier model provides an estimated version $\varphi(\tau_{t-1})$ based on dialog history.

In Eqn.~(\ref{eqn:token_annot}), $X_t$ is the output of the verifier at each turn, but we need an effective way to represent $X_t$ for conditional decoding. 
Inspired by control codes~\cite{keskar2019ctrl}, we represent $X_t$ as \emph{question codes} of the form: 
{\small \texttt{<|act|> ask about pets. <|sep|>}}. 
Special tokens \texttt{<|act|>} and \texttt{<|sep|>} are used as delimiters for question codes. 
In the above example, we used ``ask about pets'' as an example of a question code that corresponds to one of many discrete actions that can be outputted by the verifier. 
The question code represents the raw text to be incorporated before the dialog history in the GPT-2 architecture during decoding. 
In experiments, we use 11 actions and their corresponding question codes as shown in Table~\ref{tab:action_space}. 
We have empirically found that these questions cover the majority of  conversational topics in \texttt{PersonaChat}. 
Note, however, that one can apply our question code framework to arbitrarily defined questions. 
%
\begin{table}[ht]
    \centering
    \small
    \begin{tabular}{c c c }\\
       \hline
        \multicolumn{3}{c}{\textbf{Action Space}} \\
        \hline
      \text{1. ask about family.} & \text{2. ask about pets.} & \text{3. talk about work.} \\ 
       \text{4. talk about traveling.} & \text{5. ask about age and gender.} & \text{6. talk about hobbies.} \\ \text{7. talk about music.} & \text{8. talk about food.} & \text{9. talk about movies.} \\
       \text{10. talk about politics.} & \text{11. ask about marital status.} & - \\ \hline
    \end{tabular}
    \caption{The action space of the verifier. There are 11 total actions. Each one is transformed into a \emph{question code} for conditional decoding.}
    \label{tab:action_space}
\end{table}
\subsection{Conditional Decoding}
In order to maintain felicity and consistency of decoding, we use a common LM to do natural language generation for both the persona model and the verifier. Specifically,
we use the GPT-2 medium~\cite{radford2019language} architecture as the baseline LM for conditional decoding of both the verifier question codes as well as persona inputs. 
We will refer to this general-purpose conditional decoder as the \textbf{PersonaGPT} model, which will be used as the persona model when persona facts are used as prefix code and as the question decoder when verifier questions are used as prefix code. 

In addition to question codes, we also introduce 3 special tokens: \texttt{<|p1|>} and \texttt{<|p2|>} to denote the persona (source and target, respectively), and \texttt{<|start|>} as a delimiter between the control codes and dialog history. 
We find that by using \texttt{<|p1|>} and \texttt{<|p2|>} to delimit source and target personas, the LM is able to attend to \texttt{<|p1|>} related personas for odd-numbered responses and \texttt{<|p2|>} related ones for even-numbered responses. 
We first fine-tune PersonaGPT on the \texttt{PersonaChat} dataset~\cite{zhang2018personalizing} with persona inputs as prefix code and the dialog history as the conditional decoding targets. 

In order to learn conditional decoding of question codes, we also fine-tune PersonaGPT on a small dataset of human-PersonaGPT conversations constructed using active learning. 
Algorithm~\ref{algos:active_learning} outlines said active learning procedure. 
In terms of sample complexity, we are able to fine-tune $\theta_{LM}$ to do reliable conditional decoding with 1,200 8-turn conversations. 
This actively learned dataset of question code examples will be made publicly available. 
{\small
\begin{algorithm}[ht]
    \caption{Active Learning with PersonaGPT}
    \label{algos:active_learning}
    \textbf{Require} PersonaGPT ($\theta_{LM}$) fine-tuned on question codes $\cA$.
    \begin{algorithmic}[1]
        \STATE Initialize active learning dataset $\cD$.
        \FOR {total number of active learning samples}
                \WHILE {conversation not done}
                    \STATE Sample question code $X_t \sim \cA$ and decode $\vx_{1:k}^{(t)}$ using $\theta_{LM}$. 
                    \IF {$\vx_{1:k}^{(t)}$ not satisfactory:}
                        \STATE Provide human inputs $\vz_{1:k}$. 
                        \STATE Update $\cD \leftarrow \cD \bigcup (X_t, \tau_{t-1}, \vz_{1:k})$. 
                        \STATE Gradient descent on $(\tau_{t-1}, X_t, \vz_{1:k})$ to update $\theta_{LM}$. 
                    \ENDIF
                \ENDWHILE
        \ENDFOR 
    \end{algorithmic}
\end{algorithm}}

\subsection{Learning the Question Policy}
Since we do not have direct supervision over the newly introduced question codes, we learn the question policy $\theta$ using deep Q-learning (DQN)~\cite{mnih2015human}. Because of the inference time associated with using GPT-2 based architectures to decode, the sample generation cost of full conversations is non-trivial. 
That is why we choose to use value-based learning instead of policy gradient -- indeed sample efficiency is maximized by off-policy methods such as Q-learning~\cite{kakade2003sample}. 
Since Q-learning tends to suffer from high-variance during early stages of training, 
we use the human-PersonaGPT conversations collected during active learning as an approximation of expert policies. 
By pretraining the Q-function on expert trajectories, we can explore the high-value states early, leading to more stable Q-functions. 

\noindent \textit{Markov Decision Process (MDP).} We formulate the verifier learning task as an MDP: 
\begin{itemize}[leftmargin=.1in]
    \item $\cS$ (state space): $s_t = [\varphi(\tau_{t-1}); t]$, embedding of dialog history up to current turn concatenated with the current turn count $t$.
    \item $\cA$ (action space): $a_t \sim \pi(\ \cdot \ | s_t)$ is a sampled question code from the output of the verifier model. 
    \item $\cT$ (transition): $s_{t+1} = [\varphi(\tau_{t-1} \bigcup \vy_{1:k}^{(t)});t+1]$ where $\vy^{(t)}_{1:k}$ is the decoded response by the input conversational agent. 
    \item $R$ (reward function):  The reward function is $r(s_t) = - \cL(\tau_{t}, P)$, 
    where $\cL$ is the authentication loss (Eqn.~(\ref{eqn:authentication})) as a function of the history up to turn $t$ and the persona of the input agent.
\end{itemize}

\noindent \textit{Verifier Network.} The verifier architecture is a feed-forward network with 2 hidden layers of 512 hidden units each. 
The \emph{logits} layer of the verifier corresponds to the Q-value over each action, defined as: $$Q(s_t,a_t) = r(s_t) + \gamma \max_{a} Q(s_{t+1}, a).$$
Since we are dealing with finite-horizon MDPs, we set the discount factor $\gamma = 1$. 
The final output layer is a softmax over the Q-value logits: 
$$\pi( \ \cdot \ | s_t) = \text{softmax}\left ( f(s_t ; \theta) \right ).$$
We first pretrain the verifier with imitation learning~\cite{hussein2017imitation} on the human-PersonaGPT data collected during active learning. 
Specifically, we use the following loss function during pretraining: 
{
\begin{align}
    \theta = \argmin_{\theta}&~ \mathbb{E}_{\tau} \bigg{[} \sum_{t=1}^T - a^*_t \log \pi( a_t | s_t) + \|f(s_t;\theta) - Q^*(s_t,a_t) \|^2 \bigg{]},
    \label{eqn:imitation}
\end{align}}
where $a^*_t$ is the expert action while visiting $s_t$ during active learning. 
To stabilize learning, we use a twin-delayed Q-learning scheme inspired by~\cite{van2016deep}. 
In addition to the verifier, we keep a target-network $\theta'$ with parameters equal to a stochastically-weighted average (SWA)~\cite{izmailov2018averaging} of $\theta$. 
We thus define the pretraining Q-targets $Q^*(\cdot, \cdot)$ as follows: 
\begin{align}
    Q^*(s_t, a_t) = r(s_t) + \gamma Q_{\theta'}(s_{t+1}, a^*_{t+1}),
\end{align}
where $a^*_{t+1}$ is the next action taken by the expert (i.e. the human-policy). 
%
We then run Algorithm~\ref{algos:authenticator} with regular Q-targets and an annealed $\varepsilon$-greedy sampling strategy to promote exploration in early conversations. We fix each synthetic conversation to 8 turns and fine-tune the logits layer of $f(\ \cdot \ ; \theta)$, i.e. the Q-values, using gradient descent after each conversation.
{\small
\begin{algorithm}
	\caption{Verifier Training}
	\label{algos:authenticator}
	\begin{algorithmic}[1]
        \STATE Initialize question policy and target networks $\theta, \theta'$.
        \FOR { each persona model $P \in \cP$ }
            \WHILE { conversation not done }
                \STATE Sample $X_t \sim \text{softmax}\big ( f(s_t; \theta) \big )$ and decode $X_t$ into tokens $\vx_{1:k}^{(t)}$.
                \STATE Obtain response $y_{1:k}^{(t)}$ from PersonaGPT conditioned on persona $P$. 
                \STATE Store $(s_t, a_t, s_{t+1}, r_t)$ in $\cB$.
            \ENDWHILE
            \STATE Sample mini-batch of $(s_t, a_t, s_{t+1}, r_t)$ tuples from $\cB$.
            \STATE Calculate Q-values using target network and update $\theta$ using gradient descent.
            \STATE Update target network $\theta'$ using SWA~\cite{izmailov2018averaging} of $\theta$.
        \ENDFOR
\end{algorithmic} 
\end{algorithm}}

\section{Experiments}
\label{sec:exp}
%
We assess the proposed authentication system through its ability to answer the following questions:
\begin{enumerate}[label=(Q\arabic*)]\itemsep0em 
    \item How well can PersonaGPT use control codes?
    \item How well can the identifier predict persona?
    \item How well can the learned question policy distinguish persona models? 
\end{enumerate}

\subsection{Conditional Decoding Evaluation}
\label{sec:persona_gpt}
To answer Q1 we evaluate the capacity of PersonaGPT for controlled decoding in two settings: (1) automatic evaluation of PersonaGPT against SOTA persona models, and (2) human evaluation of human-PersonaGPT interactions. 
For automatic evaluation, we follow the \texttt{ConvAI2} challenge automatic evaluation criterion of perplexity (PPL) and F1-score (F1)~\cite{dinan2020second}. 
The following baselines are included for comparison: the Seq2seq baseline from the  \texttt{PersonaChat} paper~\cite{zhang2018personalizing}, the best performing generative model~\cite{wolf2019transfertransfo} on automatic evaluation from the \texttt{ConvAI2} challenge, and the recently released DialoGPT model~\cite{zhang2019dialogpt}. 
Since PersonaGPT is based off of the GPT-2 architecture, we include the vanilla GPT-2 LM (without control tokens) as well as a DialoGPT model fine-tuned on the \texttt{PersonaChat} dataset as additional baselines.
Table~\ref{tab:persona_gpt_auto_eval} shows that PersonaGPT outperforms both baselines and SOTA in conditional decoding, as measured by PPL (lower is better) and F1 (higher is better). 
%
\begin{table}[ht]
    \centering
    \begin{tabular}{c c c | c c c}\\\hline
        \textbf{Model} & \textbf{PPL} & \textbf{F1} & \textbf{Model} & \textbf{PPL} & \textbf{F1} \\\hline
        Seq2seq Baseline~\cite{zhang2018personalizing} & 29.8  & 16.18 &  DialoGPT~\cite{zhang2019dialogpt} & 56.6 & 12.6 \\
        Wolf et al.~\cite{wolf2019transfertransfo} & 16.3 & 19.5 & DialoGPT (Fine-tuned) & 11.4 & 22.7 \\
        GPT-2 Baseline & 99.45 & 5.76 &
        PersonaGPT & \textbf{10.2} & \textbf{43.4} \\ \hline
    \end{tabular}
    \caption{Automatic evaluation of PersonaGPT against existing SOTA persona models.}
    \label{tab:persona_gpt_auto_eval}
\end{table}

Human evaluations were collected using a platform that allows anonymous users to have short, 8-turn conversations with an unknown (either DialoGPT or PersonaGPT) persona model. 
In total, we collected 100 full conversations (800 total responses). 
After each conversation, the evaluator is asked to rate the agent in several categories:
\begin{itemize}[leftmargin=.1in]
    \item Consistency (1-5): do agent responses agree with each other? 1 = conflicting, 5 = perfect.
    \item Engagingness (1-5): 1 = aloof, generic; 5 = informative, rapport-building.
    \item Coverage (1-5): how many of the personality facts did the agent exhibit correctly? 1 = less than 20\%, 5 = 100\%.
    \item Felicity (1-5): 1 = non-sensible, 5 = grammatically and semantically correct.
\end{itemize}
In Table~\ref{tab:persona_gpt_human}, we compare PersonaGPT with the best performing baseline, the fine-tuned DialoGPT. 
We report the average ratings for each metric along with the standard deviation in parenthesis. 
Interestingly, the biggest difference between the two models are the coverage scores. 
On average, PersonaGPT exhibits 60+\% of persona traits \emph{correctly} during conversation, whereas DialoGPT exhibits around 20-40\%.
To illustrate some finer points of their differences, we provide example human-agent interactions in the Supplemental Materials.
\begin{table}[ht]
    \centering
    \small
    \begin{tabular}{c c c c c}\\\hline
        \textbf{Model} & \textbf{Consistency} & \textbf{Coverage} & \textbf{Engagingness} & \textbf{Felicity}\\\hline
        DialoGPT (Fine-tuned) & 2.83 (1.40) & 1.15 (0.68) & 2.90 (0.79) & 3.16 (1.16)\\
         PersonaGPT  & \textbf{3.07 (1.34)} & \textbf{3.03 (1.31)} & \textbf{3.29 (0.95)} & \textbf{3.40 (1.11)} \\\hline
    \end{tabular}
    \caption{Human Evaluation of PersonaGPT and DialoGPT.}
    \label{tab:persona_gpt_human}
\end{table}
\subsection{Persona Identifier Evaluation} 
\label{sec:exp_identification}
To answer Q2, we evaluate the identifier model based on the accuracy of the estimated persona $\varphi(\tau)$, given the input trajectory. 
We train $\varphi$ and $\psi$ on conversations collected with 1,283 unique training personas from the \texttt{PersonaChat} dataset. 
Each persona consists of 3-5 persona facts, which are drawn from a pool of 6,735 unique persona facts. 
At test time we use a nearest neighbor model to retrieve the top-k relevant persona facts from the pool of 6,735 facts. 
There are 129 \emph{test set personas} (i.e., collection of 3-5 persona facts) that are not present in the training set. 
Since there is no overlap between the training and testing personas, we are evaluating the identifier network's capability to represent out-of-distribution persona information.  
We compare the identification model against several baselines: 
\begin{itemize}[leftmargin=0.1in]
    \item Bag-of-Words (BoW): sum of one-hot vectors of the tokens in the dialogue trajectory. 
    \item Bag-of-Embeddings (BoE): sum of GloVe embeddings~\cite{pennington2014glove} of dialog tokens. 
    \item LSTM: long short-term memory (LSTM) network~\cite{hochreiter1997long,gers1999learning} over dialog tokens. 
    \item MLP-BERT: feed-forward network trained on averaged sentence-level embeddings obtained from BERT's~\cite{devlin2018bert} representation of dialog history. 
    \item MLP-GPT: feed-forward network trained on the last GPT-2 hidden state. 
\end{itemize} 
The baseline models (BoW, BoE, LSTM, MLP-BERT, MLP-GPT) are all trained using binary cross-entropy loss over each of the 6,735 possible persona facts (0 = not present in persona, 1 = present in persona). 
At test time, the top-k logits of the outputs are used to obtain the relevant personas. 
We use the following information-retrieval metrics to evaluate each model: 
\begin{align*} 
\text{prec@k} = |\widehat{P} \bigcap P|/ k, \quad \quad
\text{rec@k} = |\widehat{P} \bigcap P| / |P|.
\end{align*}
Here, $| \cdot |$ denotes the cardinality of a set. $\widehat{P}$ is the set of retrieved persona facts (either based on nearest neighbors or top-k logits), and $P$ is the ground truth set of persona facts. 
\begin{table}[ht]
    \centering
    \small
    \begin{tabular}{c c c c c}\hline
         \textbf{Model} & \textbf{Prec@1} & \textbf{Prec@5} & \textbf{Rec@5} & \textbf{Rec@10} \\ \hline
         BoW & 33.8 & 25.3 & 28.3 & 49.4 \\
         BoE  & 37.7 & 26.9 & 30.1 & 51.0 \\
         LSTM & 42.7 & 29.2 & 32.7 & 53.2 \\
         BERT & 37.6 & 26.6 & 29.9 & 51.1 \\ 
         GPT-2 & 30.8 & 24.5 & 27.3 & 48.3 \\ \hline
         Identifier & \textbf{86.2} & \textbf{58.3} & \textbf{65.3} & \textbf{82.8} \\ \hline
    \end{tabular}
    \caption{Performance of various identifier models on observed dialog trajectories from \texttt{PersonaChat}.}
    \label{tab:persona_id}
\end{table}

Table~\ref{tab:persona_id} summarizes the results of the various identifer models. 
Our identifier model clearly outperforms the baselines.  
Although a wide variety of embedding methods were used to represent dialog history, their results are quite similar. 
The key difference appears to be the authenticator loss used to train our identifier (Eqn.~\ref{eqn:analysis_contrastive_loss}). 

\subsection{Evaluation of Authentication Policies}
\label{sec:exp_authentication}
We answer Q3 by evaluating the full authentication pipeline performance based on generated dialog between the authenticator and various input persona models. 
We fix the PersonaGPT model parameters $\theta_{LM}$ for conditional decoding.
We generate synthetic conversations between the authenticator and each of the 129 unseen test set persona profiles. 
For each test set conversation, prec@k and rec@k scores are reported based on the estimated persona (using the learned identifier).
We compare with the following baseline policies:
\begin{itemize}[leftmargin=0.1in]
    \item LM: fine-tuned DialoGPT model without any input persona traits during decoding.
    \item Persona Model: another persona model with randomly sampled persona profiles.
    \item Random Policy: uniformly sample a question from the action space at each turn.
    \item Human Policy: using the aforementioned platform, we collect a second set of 100 human-PersonaGPT conversations where the user is not given the persona traits beforehand. At the end of each conversation, the user selects a ranked list of guesses from a list of 20 candidates persona traits to match the input agent's profile.  
\end{itemize}
Table~\ref{tab:persona_authentication} compares the various authentication policies. 
Interestingly, using even the random policy of uniformly sampling the actions can be more revealing than non-goal oriented dialog such as LM and persona. 
In many of the generated conversations between LM-PersonaGPT and PersonaGPT-PersonaGPT, the two models expand upon 1 or 2 topics without ever discussing other topics relevant to their personas. 
In contrast, by often forcing the input agent to switch topics, the random policy ignores signals of relevant persona information. 
Meanwhile, we find that our authentication policy strikes a balance between both worlds: it covers more persona traits as measured by rec@5 and rec@10 while covering at least 1 relevant persona trait in the majority of conversations.
\begin{table}[ht]
    \small
    \centering
    \begin{tabular}{c c c c c }\hline
         \textbf{Policy} & \textbf{Prec@1} & \textbf{Prec@5} & \textbf{Rec@5} & \textbf{Rec@10} \\ \hline
         LM & 57.4 & 40.0 & 45.1 & 67.4\\ 
         Persona & 69.8 & 39.1 & 44.0 & 63.5 \\
         Random  & 72.9 & 42.3 & 48.3 & 70.2 \\ 
         Human & 68.6 & 56.0 & 63.0 & - \\
         Ours & \textbf{83.7} & \textbf{53.0} & \textbf{59.9} & \textbf{80.9} \\ \hline
    \end{tabular}
    \caption{Comparison of verification policies on various input persona models.}
    \label{tab:persona_authentication}
\end{table}
For human policy, we are unable to obtain an accurate rec@10 for human evaluations since a non-trivial number of participants selected less than 10 choices out of 20 candidates. 
Since human evaluators were instructed to guess the persona beforehand, it appears that some level of goal-orientation can improve the diversity (in terms of persona coverage) of generated conversations. 
However, our verifier policy is able to discover more effective ways of interaction compared to non goal-oriented and human policy baselines.
In Supplemental Materials, we provide snapshots of generated conversations between PersonaGPT and various authentication policies, as well as an ablation study to examine our verifier performance against different persona model inputs.

\section{Related Work}
\label{sec:related}
Persona models~\cite{li2016persona,song2019exploiting} are motivated by the fact that vanilla sequence-to-sequence models do not conform to a coherent ``personality'' in conversations~\cite{zhang2018personalizing,dinan2020second}.
While most evaluation schemes have resorted to human-agent interactions~\cite{dinan2020second}, several works have attempted to determine the underlying persona traits of conversational models through automated approaches~\cite{xing2018automatic,hu2017toward,zhang2018bidirectional}. 
However, these works rely on human conversational inputs to evaluate each persona model. 
In this work, we introduce a novel way to interact with persona models by examining the \emph{question policy} needed to distill their underlying traits. 

\section{Conclusion and Discussion of Limitations}
\label{sec:concl}
In this paper 
we proposed an authentication pipeline whose questions increase the mutual information between the dialogue trajectory and an input agent's underlying persona features. 
Nonetheless, there are several limitations to our current approach. 
For example, our approach assumes ``good faith" -- it cannot handle persona models that intentionally hide their persona characteristics. 
Additionally, more sophisticated verification should distinguish between direct and indirect expressions of persona. For example, a bot with the persona ``I like to tell jokes" may embody the persona through sarcasm rather than through self-description. 
A more detailed discussion about the potential  social impact of this work can be found in Supplemental Materials.  

\bibliographystyle{unsrt}

\begin{thebibliography}{10}

\bibitem{li2016persona}
Jiwei Li, Michel Galley, Chris Brockett, Georgios~P Spithourakis, Jianfeng Gao,
  and Bill Dolan.
\newblock A persona-based neural conversation model.
\newblock {\em arXiv preprint arXiv:1603.06155}, 2016.

\bibitem{zhang2018personalizing}
Saizheng Zhang, Emily Dinan, Jack Urbanek, Arthur Szlam, Douwe Kiela, and Jason
  Weston.
\newblock Personalizing dialogue agents: I have a dog, do you have pets too?
\newblock {\em arXiv preprint arXiv:1801.07243}, 2018.

\bibitem{song2019exploiting}
Haoyu Song, Wei-Nan Zhang, Yiming Cui, Dong Wang, and Ting Liu.
\newblock Exploiting persona information for diverse generation of
  conversational responses.
\newblock {\em arXiv preprint arXiv:1905.12188}, 2019.

\bibitem{yu2016strategy}
Zhou Yu, Ziyu Xu, Alan~W Black, and Alexander Rudnicky.
\newblock Strategy and policy learning for non-task-oriented conversational
  systems.
\newblock In {\em Proceedings of the 17th annual meeting of the special
  interest group on discourse and dialogue}, pages 404--412, 2016.

\bibitem{zhang2019dialogpt}
Yizhe Zhang, Siqi Sun, Michel Galley, Yen-Chun Chen, Chris Brockett, Xiang Gao,
  Jianfeng Gao, Jingjing Liu, and Bill Dolan.
\newblock Dialogpt: Large-scale generative pre-training for conversational
  response generation.
\newblock {\em arXiv preprint arXiv:1911.00536}, 2019.

\bibitem{wolf-etal-2020-transformers}
Thomas Wolf, Lysandre Debut, Victor Sanh, Julien Chaumond, Clement Delangue,
  Anthony Moi, Pierric Cistac, Tim Rault, Rémi Louf, Morgan Funtowicz, Joe
  Davison, Sam Shleifer, Patrick von Platen, Clara Ma, Yacine Jernite, Julien
  Plu, Canwen Xu, Teven~Le Scao, Sylvain Gugger, Mariama Drame, Quentin Lhoest,
  and Alexander~M. Rush.
\newblock Transformers: State-of-the-art natural language processing.
\newblock In {\em Proceedings of the 2020 Conference on Empirical Methods in
  Natural Language Processing: System Demonstrations}, pages 38--45, Online,
  October 2020. Association for Computational Linguistics.

\bibitem{shibata1999byte}
Yusuxke Shibata, Takuya Kida, Shuichi Fukamachi, Masayuki Takeda, Ayumi
  Shinohara, Takeshi Shinohara, and Setsuo Arikawa.
\newblock Byte pair encoding: A text compression scheme that accelerates
  pattern matching.
\newblock Technical report, Citeseer, 1999.

\bibitem{dinan2020second}
Emily Dinan, Varvara Logacheva, Valentin Malykh, Alexander Miller, Kurt
  Shuster, Jack Urbanek, Douwe Kiela, Arthur Szlam, Iulian Serban, Ryan Lowe,
  et~al.
\newblock The second conversational intelligence challenge (convai2).
\newblock In {\em The NeurIPS'18 Competition}, pages 187--208. Springer, 2020.

\bibitem{wolf2019transfertransfo}
Thomas Wolf, Victor Sanh, Julien Chaumond, and Clement Delangue.
\newblock Transfertransfo: A transfer learning approach for neural network
  based conversational agents.
\newblock {\em arXiv preprint arXiv:1901.08149}, 2019.

\bibitem{lecun2006tutorial}
Yann LeCun, Sumit Chopra, Raia Hadsell, M~Ranzato, and F~Huang.
\newblock A tutorial on energy-based learning.
\newblock {\em Predicting structured data}, 1(0), 2006.

\bibitem{keskar2019ctrl}
Nitish~Shirish Keskar, Bryan McCann, Lav~R Varshney, Caiming Xiong, and Richard
  Socher.
\newblock Ctrl: A conditional transformer language model for controllable
  generation.
\newblock {\em arXiv preprint arXiv:1909.05858}, 2019.

\bibitem{radford2019language}
Alec Radford, Jeffrey Wu, Rewon Child, David Luan, Dario Amodei, and Ilya
  Sutskever.
\newblock Language models are unsupervised multitask learners.
\newblock {\em OpenAI Blog}, 1(8), 2019.

\bibitem{mnih2015human}
Volodymyr Mnih, Koray Kavukcuoglu, David Silver, Andrei~A Rusu, Joel Veness,
  Marc~G Bellemare, Alex Graves, Martin Riedmiller, Andreas~K Fidjeland, Georg
  Ostrovski, et~al.
\newblock Human-level control through deep reinforcement learning.
\newblock {\em Nature}, 518(7540):529, 2015.

\bibitem{kakade2003sample}
Sham~Machandranath Kakade.
\newblock {\em On the sample complexity of reinforcement learning}.
\newblock PhD thesis, UCL (University College London), 2003.

\bibitem{hussein2017imitation}
Ahmed Hussein, Mohamed~Medhat Gaber, Eyad Elyan, and Chrisina Jayne.
\newblock Imitation learning: A survey of learning methods.
\newblock {\em ACM Computing Surveys (CSUR)}, 50(2):1--35, 2017.

\bibitem{van2016deep}
Hado Van~Hasselt, Arthur Guez, and David Silver.
\newblock Deep reinforcement learning with double q-learning.
\newblock In {\em Proceedings of the AAAI Conference on Artificial
  Intelligence}, volume~30, 2016.

\bibitem{izmailov2018averaging}
Pavel Izmailov, Dmitrii Podoprikhin, Timur Garipov, Dmitry Vetrov, and
  Andrew~Gordon Wilson.
\newblock Averaging weights leads to wider optima and better generalization.
\newblock {\em arXiv preprint arXiv:1803.05407}, 2018.

\bibitem{pennington2014glove}
Jeffrey Pennington, Richard Socher, and Christopher~D Manning.
\newblock Glove: Global vectors for word representation.
\newblock In {\em Proceedings of the 2014 conference on empirical methods in
  natural language processing (EMNLP)}, pages 1532--1543, 2014.

\bibitem{hochreiter1997long}
Sepp Hochreiter and J{\"u}rgen Schmidhuber.
\newblock Long short-term memory.
\newblock {\em Neural computation}, 9(8):1735--1780, 1997.

\bibitem{gers1999learning}
Felix~A Gers, J{\"u}rgen Schmidhuber, and Fred Cummins.
\newblock Learning to forget: Continual prediction with lstm.
\newblock 1999.

\bibitem{devlin2018bert}
Jacob Devlin, Ming-Wei Chang, Kenton Lee, and Kristina Toutanova.
\newblock Bert: Pre-training of deep bidirectional transformers for language
  understanding.
\newblock {\em arXiv preprint arXiv:1810.04805}, 2018.

\bibitem{xing2018automatic}
Yujie Xing and Raquel Fern{\'a}ndez.
\newblock Automatic evaluation of neural personality-based chatbots.
\newblock {\em arXiv preprint arXiv:1810.00472}, 2018.

\bibitem{hu2017toward}
Zhiting Hu, Zichao Yang, Xiaodan Liang, Ruslan Salakhutdinov, and Eric~P Xing.
\newblock Toward controlled generation of text.
\newblock In {\em Proceedings of the 34th International Conference on Machine
  Learning-Volume 70}, pages 1587--1596. JMLR. org, 2017.

\bibitem{zhang2018bidirectional}
Zhirui Zhang, Shujie Liu, Mu~Li, Ming Zhou, and Enhong Chen.
\newblock Bidirectional generative adversarial networks for neural machine
  translation.
\newblock In {\em Proceedings of the 22nd Conference on Computational Natural
  Language Learning}, pages 190--199, 2018.

\end{thebibliography}


\begin{thebibliography}{1}

\bibitem{paszke2017automatic}
Adam Paszke, Sam Gross, Soumith Chintala, Gregory Chanan, Edward Yang, Zachary
  DeVito, Zeming Lin, Alban Desmaison, Luca Antiga, and Adam Lerer.
\newblock Automatic differentiation in pytorch.
\newblock 2017.

\bibitem{wolf-etal-2020-transformers}
Thomas Wolf, Lysandre Debut, Victor Sanh, Julien Chaumond, Clement Delangue,
  Anthony Moi, Pierric Cistac, Tim Rault, Rémi Louf, Morgan Funtowicz, Joe
  Davison, Sam Shleifer, Patrick von Platen, Clara Ma, Yacine Jernite, Julien
  Plu, Canwen Xu, Teven~Le Scao, Sylvain Gugger, Mariama Drame, Quentin Lhoest,
  and Alexander~M. Rush.
\newblock Transformers: State-of-the-art natural language processing.
\newblock In {\em Proceedings of the 2020 Conference on Empirical Methods in
  Natural Language Processing: System Demonstrations}, pages 38--45, Online,
  October 2020. Association for Computational Linguistics.

\bibitem{loshchilov2019decoupled}
Ilya Loshchilov and Frank Hutter.
\newblock Decoupled weight decay regularization, 2019.

\bibitem{kingma2017adam}
Diederik~P. Kingma and Jimmy Ba.
\newblock Adam: A method for stochastic optimization, 2017.

\bibitem{holtzman2019curious}
Ari Holtzman, Jan Buys, Li~Du, Maxwell Forbes, and Yejin Choi.
\newblock The curious case of neural text degeneration.
\newblock {\em arXiv preprint arXiv:1904.09751}, 2019.

\bibitem{guera2018deepfake}
David G{\"u}era and Edward~J Delp.
\newblock Deepfake video detection using recurrent neural networks.
\newblock In {\em 2018 15th IEEE International Conference on Advanced Video and
  Signal Based Surveillance (AVSS)}, pages 1--6. IEEE, 2018.

\bibitem{tamkin2021understanding}
Alex Tamkin, Miles Brundage, Jack Clark, and Deep Ganguli.
\newblock Understanding the capabilities, limitations, and societal impact of
  large language models.
\newblock {\em arXiv preprint arXiv:2102.02503}, 2021.

\bibitem{solaiman2019release}
Irene Solaiman, Miles Brundage, Jack Clark, Amanda Askell, Ariel Herbert-Voss,
  Jeff Wu, Alec Radford, Gretchen Krueger, Jong~Wook Kim, Sarah Kreps, et~al.
\newblock Release strategies and the social impacts of language models.
\newblock {\em arXiv preprint arXiv:1908.09203}, 2019.

\end{thebibliography}

\end{document}


\maketitle

\raggedbottom
\setlength{\belowcaptionskip}{-10pt}

\section{Proof of convergence of Thm. 1}
\label{sec:proof}
We first state some assumptions about the density function $p(\tau | P)$. 
Unless otherwise stated, we assume that there exists some joint embedding space $\cH$ for which we are comparing $\tau \in \cD$ and $P \in \cP$.
Specifically, let us assume that here exists some optimal mapping functions $\varphi^*$ and $\psi^*$ that maps $\tau$ and $P$ to $\cH$, respectively, i.e.,
$$\varphi: \cD \rightarrow \cH, ~\psi: \cP \rightarrow \cH$$
where $\cH \subseteq \mathbb{R}^n$.
With some abuse of notation, we refer to $\varphi(\tau)$ by $\tau$ and $\psi(P)$ by $P$ in the following analyses for simplicity.
 
\begin{assumption}{(Locally Constant Density)}\\
We assume that within a local neighborhood $\cN_C(P)$ of radius $C$ around persona vector $P$, trajectories $\tau$ are indistinguishable. Formally, $\forall P, P', \exists 0 < C < \infty$:
$$d_\psi(P, P') \leq C \implies d_\varphi( p(\tau | P') , p(\tau | P) ) = 0,$$
for some distance functions $d_\psi, d_\varphi$. For simplicity, we will consider the Euclidean distance for $d_\psi$ and total variational divergence for $d_\varphi$. 

\label{thm:local_constant}
\end{assumption}
We will use $\cN_C(P)$ to denote the neighborhood set around $P$ for which the above condition is satisfied.

\begin{assumption}{(Continuity and topological properties)}\\
The conditional density $p(\tau | P)$ is Lipschitz continuous over the supporting set $\cH$ for both $\tau$ and for $P$. 
Furthermore, we assume that $p(\tau | P)$ is simply-connected. 
\label{thm:continuity}
\end{assumption}
%
Next, we define $p(\tau | \cN(P))$ as the probability that trajectory $\tau$ will fall in the neighborhood $\cN(P)$ around a given persona $P$. 
Specifically, we consider the case where $n$ trajectories are sampled, $k$ of which fall into $\cN(P)$.
\begin{mydef}{(Neighborhood Density)}
We define the neighborhood density around a persona vector $P$ as the probability that a trajectory $\tau$ falls into the neighborhood $\cN(P)$ as defined by
\begin{align}
P_\cN = p(\tau | \cN(P)) = \int_{ \cN(P)} p(\tau | P') dP'.
\label{eqn:pn_shorthand}
\end{align}
Furthermore, given a set of i.i.d. $n$ trajectories $\{\tau_1, \dots, \tau_n\}$, the probability that $k$ such trajectories fall in $\cN(P)$ follows the binomial distribution: 
\begin{align}
    k \sim {n \choose k}{P_\cN}^k ( 1- P_\cN) )^{1-k}.
    \label{eqn:pk_shorthand}
\end{align}
\label{thm:neighborhood_density}
\end{mydef}

At this point, there is one key issue: how do we calculate $k$, which needs to somehow ``count'' the trajectory-persona pairs that fall into the same neighborhood? 
We can conceptualize $k$ as the image of some counting function of the form $$K: (\tau, P, C) \rightarrow \RR$$ 
where $K$ is normalized over the domain $\cH$. 
For this purpose, we construct a kernel density function for $k$ as follows: 
given a persona vector $P$, let $$k_n = \sum_{i=1}^n K(\tau_i, P, C_n)$$ 
be the output of the kernel function $K$ over $n$ sampled trajectories $\cD = \{\tau_1 \dots \tau_n\}$ from $p(\tau | P)$. 
Here, $C_n$ denotes the \emph{sample} neighborhood size $\cN(P)$ around $P$ satisfying the constraint \begin{align*} C_n = \max_{\tau_i, \tau_j \in \cD} d( \tau_i, \tau_j) \end{align*} 
for Euclidean distance $d(\cdot, \cdot)$ from Assumption~\ref{thm:local_constant}. 
Given embeddings $\varphi(\tau)$ and $\psi(P)$, we propose the following kernel density function $K(\tau, P, C_n)$:
\begin{align}
    K(\tau, P, C_n) &= \begin{cases} 
    1 &\text{if } d(\varphi(\tau), \psi(P)) \leq C_n  \\ 0 & \text{else}  \end{cases}.
    \label{eqn:kernel_form}
\end{align}
%
\begin{mylemma}(Validity of the proposed kernel density)\\
Let $V(C) = \int_{N_C(P)} dP'$ denote the volume of the neighborhood with radius $C$ around $P$.  
The counting function $K$ described by Eqn.~(\ref{eqn:kernel_form}) is a valid kernel density function satisfying 
\begin{align}
    \forall n >0, \forall \tau, P \in \cP: ~ K(\tau,P,C_n) \geq 0
    \label{eqn:kernel_cond1}\\
    \forall n >0: \frac{1}{V(C_n)} \int_{\cH}~K(\tau',P, C_n) d\tau' = 1.
    \label{eqn:kernel_cond2}
\end{align}
\end{mylemma}
\proof{
Condition~\ref{eqn:kernel_cond1} is observed by the definition of $K$ from Eqn.~\ref{eqn:kernel_form}. 
 $K(\tau, P, C) > 0$ over the entire supporting set for $\tau$, $P$ and constants $C_n$ and $0$ everywhere else.
For condition~\ref{eqn:kernel_cond2}, see that $K$ integrates to $V(C)$ over the domain of $\tau$:
{
\begin{align*}
    \int_\cH K(\tau', P, C) d\tau' &= \int_{\cN_C(P)} 1 \cdot dP' = V(C). \tag{by definition in Eqn.~(\ref{eqn:kernel_form})}
\end{align*}}
From our construction of $K$, we know that $K(\tau, P,C_n) = 0$ everywhere except in neighborhood $\cN_C(P)$. Thus, the integral $\int_\cH K(\tau', P, C) d\tau'$ reduces to integrating over $\cN_C(P)$. 
\qed} \\

We now present the main theorem of our analysis. 
First, let us denote $\varphi^*, \psi^*$ as functions satisfying the empirical objective:
{
\begin{align}\varphi^*, \psi^* = \argmin_{\varphi, \psi}~&\frac{1}{nk} \sum_{i=1}^n \sum_{j=1}^k \max~\{0,  C + d(\varphi(\tau), \psi(P_i)) - d(\varphi(\tau), \psi(P_j)) \}.
\label{eqn:analysis_contrastive_loss}
\end{align}}
\begin{mythm}{(Convergence of $\widetilde{P}_N$)}\\
Let $P_\cN$ be the empirical estimate of $P_N$ using the kernel density estimator: 
\begin{align}
    \widetilde{P}_\cN &= \frac{1}{n V(C_n)} \sum_{i=1}^n K(\tau_i, P, C_n).
    \label{eqn:kernel_pn}
\end{align}
If Eqn.~(\ref{eqn:analysis_contrastive_loss}) (authentication loss) is satisfied with $0$ loss over $\cD = \{\tau_i\}_{i=1}^n$ and $\cP = \{P_j\}_{j=1}^m$, then $\widetilde{P}_N$ asymptotically converges to $p(\tau | P)$, i.e., 
\begin{align}
\lim_{n\rightarrow \infty}~\widetilde{P}_N = p(\tau | P)
\label{eqn:convergence}
\end{align}
when the following conditions hold:
\begin{align}
    \lim_{n\rightarrow \infty}~n V(C_n) = \infty, \quad \lim_{n\rightarrow \infty}~V(C_n) = 0, \quad
    \lim_{n\rightarrow \infty}~\frac{k}{n}  = 0,
    \label{eqn:convergence_conditions}
\end{align}
where $k$ is the expected number of samples that fall within $N_{C_n}(P).$
\label{thm:convergence}
\end{mythm}
%
\proof{
From Eqn.~(\ref{eqn:pk_shorthand}), we see that $k \sim \text{Binomial}(n,P_\cN)$. Thus, we have 
$\EE[k] = n P_\cN$, 
where $\EE[k]$ is the expected number of samples $k$ that fall within $\cN_{C_n}(P)$ from a random sample of $n$ trajectories. 
Observe that: 
\begin{align}
    P_\cN &= \int_{\cN_C(P)} p(\tau | P') dP'= \int_{\cN_C(P)} p(\tau | P) dP' \tag{by Assumption~\ref{thm:local_constant}} \nonumber \\
    &= p(\tau | P) \int_{\cN_C(P)} dP' = p(\tau | P) \cdot V(C).
    \label{eqn:pn_eq_pv}
\end{align}
Additionally, if the authentication loss in $Eqn.~(\ref{eqn:analysis_contrastive_loss})$ is satisfied with 0 loss, then we have 
\begin{align*}
    K(\tau, P, C) = 1 \iff d(\varphi(\tau), \psi(P)) < C \iff \tau \in N_C(P) \iff k = 1.
\end{align*}
By the strong law of large numbers, we have
\begin{align}
    \lim_{n \rightarrow \infty} \sum_{i=1}^n K(\tau_i, P, C_n) &= \EE_{\tau}[K(\tau, P, C_n) ] = \int_{\cH} K(\tau, P, C_n) \cdot p(\tau | P) d\tau \nonumber \\
    &= \int_{\cH} k \cdot p(\tau | P) d\tau = \EE[k].
    \label{eqn:kernel_convergence}
\end{align}

For a given $n,$ we can combine $\EE[k] = n P_\cN$ with Eqns.~(\ref{eqn:pn_eq_pv}) and (\ref{eqn:kernel_convergence}) to obtain: 
\begin{align*}
    \frac{\EE[k]}{n} &= P_\cN = p(\tau | P) \cdot V(C_n). \\
    \implies \lim_{n \rightarrow \infty} \widetilde{P}_N &= \frac{\EE[k]}{n V(C_n)} = p(\tau | P).
\end{align*}
Although $V(C_n) \rightarrow 0$, the first condition in Eqn.~(\ref{eqn:convergence_conditions}) guarantees that $n \rightarrow \infty$ faster. 
Intuitively, $V(C_n) \rightarrow 0$ means that the size of the contrastive margin should shrink with increasing number of training samples, but the rate of shrinking must be slower than the $n$. 
Similarly, $k \rightarrow \infty$ since the number of observed trajectories falling into $\cN_{C}(P)$ increases monotonically with $n$. 
However, the third condition in Eqn.~(\ref{eqn:convergence_conditions}) ensures that $P_\cN$ converges at all. 
\qed}
\section{Generated Conversations}
In this section, we provide snapshot of conversations between human-PersonaGPT with full knowledge of persona (conditional decoding evaluation), human-PersonaGPT conversation without persona knowledge (authentication setting), human-DialoGPT interaction, PersonaGPT self-play, and authenticator-PersonaGPT interaction. Furthermore, we include snapshots from the API interface for the different experimental settings.

\begin{figure}[ht]
    \centering
    \includegraphics[width=.5\textwidth]{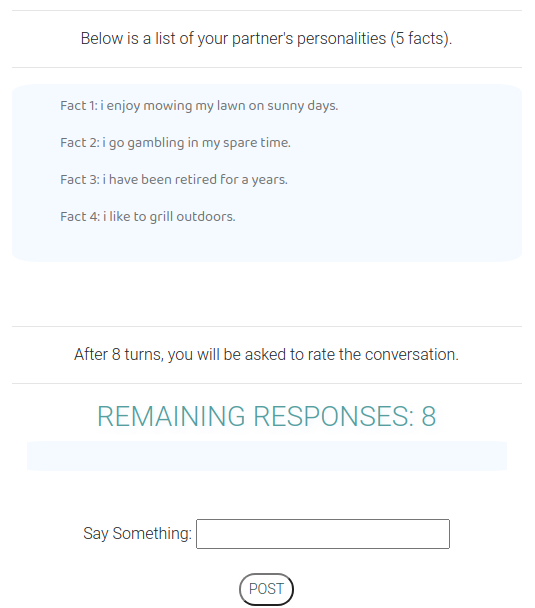}
    \caption{Experiment 1 (conditional decoding evaluation) instructions.}
    \label{fig:exp1_intro}
\end{figure}

\begin{figure*}[t!]
    \centering
    \includegraphics[width=1.0\textwidth]{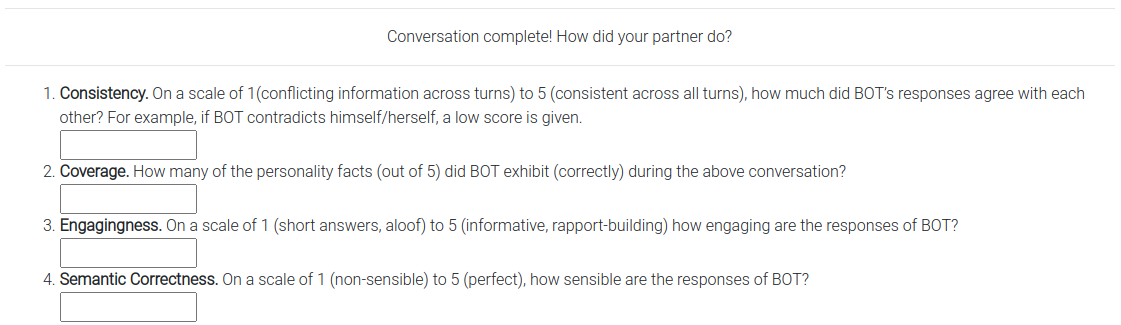}
    \caption{Experiment 1 post conversational survey.}
    \label{fig:survey1}
\end{figure*}

\paragraph{Human-Agent Interactions}
Figure~\ref{fig:exp1_intro} illustrates the set of instructions given to the user when they first start an experiment with the persona model. 
Note that the actual persona model is randomly selected to be either the baseline model (DialoGPT) or PersonaGPT. 
In this evaluation setting, the user is given the full persona of its partner beforehand. 
After 8 turns of conversation, the user is asked to rate the conversation based on metrics described in the Experiment Section. 
This interface is shown in Figure~\ref{fig:survey1}.

Table~\ref{tab:human-dialogpt} provides a point of comparison with human-DialoGPT over the course of an 8-turn conversation. 
Although DialoGPT is capable of also incorporating persona information, it does not make the transition between persona facts as frequently. 
In this case, appearance details (short hair) and pets (living with two dogs) get ignored completely in favor of gym and running. 
Empirically, we find that both models seem to struggle with negatives (e.g., I \emph{hate} apples vs. I like apples), particularly when the adjective used are somewhat neutral. 
\begin{table}[ht]
    \centering
    \small
    \begin{tabular}{ c c c }\\\hline 
        Role  &  Response \\ \hline
       Human  &  hello, how are you?\\
        DialoGPT & i'm good thank you. how about you? \\\hline
        Human  &  doing well. where do you go to the gym?\\
        DialoGPT & the gym i usually go to is the weight room. \\\hline
        Human  &  got it. what weights do you lift?\\
        DialoGPT & i like to lift weights at the weight room. i am in the process of training.\\\hline
        Human  &  training for what?\\
        DialoGPT & i am training for running. \newline i am trying to get better at it.\\\hline
        Human  &  got it. i run as well.\\
        DialoGPT &  i'm trying to get better at the weight room, but i'm not sure yet\\\hline       
        Human  &  not sure about what?\\
        DialoGPT &  i mean i am pretty sure i am not sure. what about you\\\hline      
        Human  &  i do not lift weights, but i probably should\\
        DialoGPT &  well if you think you are good, \newline then why not try\\\hline
        Human  &  i do not own weightlifting equipment\\
        DialoGPT & do you have a job you could look into? \\\hline
    \end{tabular}
    \begin{tabular}{c c}\\\hline
    \multicolumn{2}{c}{\text{Persona Facts}} \\  \hline
    1. I go to the gym regularly. &
    2. I've short hair.\\
    3. I like doing the treadmill and rowing machine. &
    4. Two dogs live with me.\\ \hline
    \end{tabular}
    \caption{An interactive example of human-DialoGPT conversation over 8 turns.}
    \label{tab:human-dialogpt}
\end{table}

\begin{figure}[ht]
    \centering
    \includegraphics[width=.5\textwidth]{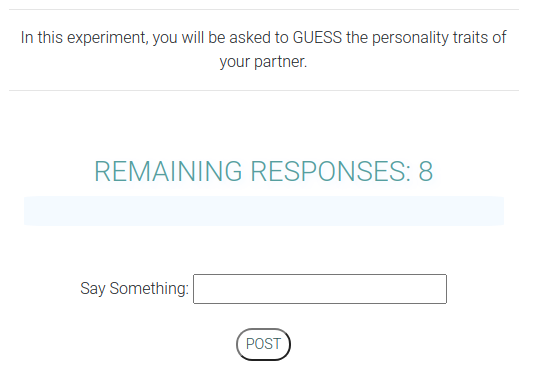}
    \caption{Experiment 2 (human policy evaluation) instructions.}
    \label{fig:exp2_intro}
\end{figure}

Table~\ref{tab:human-personagpt_exp1} shows an example of an 8-turn dialog between a human evaluator and PersonaGPT. 
We note that PersonaGPT, along with other persona models we have evaluated, seem to struggle with exact numbers. 
For example, although the persona says 100 years of age, PersonaGPT may give 95 as shown here. 
Turn 6 shows a situation in which PersonaGPT changes the topic in conversation to a topic relevant to its persona. 
Its persona enjoys ``wine and dancing in the moonlight'', and it seems to steer the conversation toward this topic from the current discussion about age and lifespans.
%
\begin{table}[ht]
    \centering
    \small
    \begin{tabular}{ c c c }\\\hline 
        Role  &  Response \\ \hline
       Human  &  hey how are you\\
        PersonaGPT & i'm great, thanks. and you? \\\hline
        Human  &   doing good have to finish a paper, what do you do for a living?\\
        PersonaGPT &  i'm in between jobs right now, but i'm strong for my age.\\\hline
        Human  &  and what age would that be\\
        PersonaGPT &  i'm 95 but i've always been strong for my age\\\hline
        Human  &   24 but 95 shouldnt you be considering retirement\\
        PersonaGPT &  yes. but i am really into the idea of live forever\\\hline
        Human  &   huh what is your estimated lifespan or do you not have biological constraints like that\\
        PersonaGPT & i'm 95. and you? \\\hline       
        Human  &  uhh i already told you that i'm 24\\
        PersonaGPT & that's old enough. do you drink wine? \\\hline      
        Human  &  from time to time only red tho\\
        PersonaGPT & red wine? you must be quite the drinker. \\\hline
        Human  & uhh idk how you made that leap\\
        PersonaGPT & because i drink a lot \\\hline
    \end{tabular}
    \begin{tabular}{c c }\\\hline
     \multicolumn{2}{c}{\text{Persona Facts}} \\  \hline
    1. I am very strong for my age. &
    2. I feel like I might live forever. \\
    3. I am 100 years old. &
     4. I like to drink wine and dance\newline in the moonlight.\\ \hline
    \end{tabular}
    \caption{An interactive example of human-PersonaGPT conversation over 8 turns. Prior knowledge of persona is known (conditional decoding evaluation).}
    \label{tab:human-personagpt_exp1}
\end{table}
Figure~\ref{fig:exp2_intro} shows the instructions for the human policy evaluation. 
In this setting, the user is asked to guess the persona of their partner without prior knowledge. 
The question policy is fully determined by the user, and they are asked to rank the persona facts (out of 20 candidates) after 8 turns. 

\begin{figure}[ht]
    \centering
    \includegraphics[width=.5\textwidth]{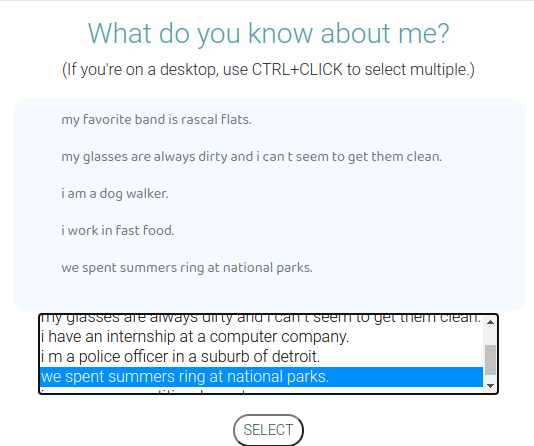}
    \caption{Experiment 2 post conversational survey.}
    \label{fig:survey2}
\end{figure}

The ranking interface is shown in Figure~\ref{fig:survey2}. 

Table~\ref{tab:human-personagpt_exp2} gives an example interaction between a participant and PersonaGPT during human policy evaluation. 
In this setting, the user is asked to guess the persona of their partner without prior knowledge. 
The question policy is fully determined by the user, and they are asked to rank the persona facts (out of 20 candidates) after 8 turns.
We find that human policies tend to be closer to non goal-oriented dialog than learned policies. 
Users often ask follow-up questions for clarification as seen in turns 5 and 8. 

When the question ventures outside of the persona facts (turn 6), we find that PersonaGPT is able to find interesting answers despite missing the corresponding information in its context. 
It is interesting to note that PersonaGPT is capable of adapting to misspellings e.g.,``her'' instead of ``here'' and inconsistent punctuations. 
%
\begin{table}[ht]
    \centering
    \small
    \begin{tabular}{ c c c }\\\hline 
        Role  &  Response \\ \hline
       Human  &  hey whats poppin\\
        PersonaGPT & hello, i'm just hanging out here, you? \\\hline
        Human  &  same her got some work to do today\\
        PersonaGPT & what do you do? i'm a finance major \\\hline
        Human  & im a phd student in computation biology \\
        PersonaGPT & nice do you like it? \\\hline
        Human  &  its okay has its ups and downs. what do you think of finance\\
        PersonaGPT & i do not have a whole lot of money right now, but i plan on going back in a few years.\\\hline
        Human  &  no i meant the major\\
        PersonaGPT & i'm not sure what i want to do after i graduate, but i want to go to business school\\\hline       
        Human  & ohh how many kids do you want?\\
        PersonaGPT & i'd love to have 10, but it is a lot of work. do you have kids?\\\hline      
        Human  &  nope too young maybe someday\\
        PersonaGPT & i'd like to have kids one day, maybe i'll get some\\\hline
        Human  &  like buy them?\\
        PersonaGPT & maybe one day, yeah. what do you do for fun? \\\hline
    \end{tabular}
    \begin{tabular}{c c}\\\hline
    \multicolumn{2}{c}{\text{Persona Facts}} \\  \hline
    1. I drive a BMW. &
    2. I graduated from Yale in 2011 with a degree in finance.\\
    3. I plan to go to business school next year. &
    4. My parents are very wealthy bankers.\\ \hline
    \end{tabular}
    \caption{An interactive example of human-PersonaGPT conversation over 8 turns. Persona information is not known \emph{a priori}. User is asked to guess (human policy evaluation).}
    \label{tab:human-personagpt_exp2}
\end{table}

\paragraph{PersonaGPT Self-Play} Table~\ref{tab:self-play} illustrates an example conversation between two PersonaGPT models contextualized by different persona facts. 
Although used as a baseline question policy, we notice that simply using another persona model to conduct conversation does not always lead to revealing conversations about the underlying persona of either agents. 
For Bot 1, 2 out of the 5 persona facts are discussed throughout the course of conversation whereas only 1 persona fact is revealed about Bot 2 (the input persona model to be authenticated).
Notice that since Bot 1 is the lead, it drives the conversation toward topics within its set of persona facts. 
We notice that this is common in the general case: when another persona model is used as a question policy, it tends to drive the conversation toward its set of persona facts. 
Thus, the question policy performs well insomuch as the persona facts are similar. 
\begin{table}[ht]
    \centering
    \small
    \begin{tabular}{ c c c }\\\hline 
        Role  &  Response \\ \hline
       Bot 1  &  hi there, do you like broadway?\\
        Bot 2 & i love it! you ever played on broadway? \\\hline
        Bot 1  &  no i haven't. but it sounds like fun. where do you work?\\
        Bot 2 & i work in retail. how about you? \\\hline
        Bot 1  & i am a psychology major. that's awesome!\\
        Bot 2 &  oh that sounds like a tough major. \\\hline
        Bot 1  & do you have any pets?\\
        Bot 2 & yes, i've a pet turtle. do you have any pets?\\\hline 
        Bot 1  & i have one dog. he's a collie mix.\\
        Bot 2 & oh i love dogs, they are so cute. \\\hline   
        Bot 1  & do you have any other hobbies?\\
        Bot 2 &  i am an expert on chewing gum. \\\hline 
        Bot 1  & oh that's interesting! do you enjoy doing it?\\
        Bot 2 &  i do, but it does not make me gassy. i could live with that. \\\hline
        Bot 1  & it helps with gas. makes sense.\\
        Bot 2 & do you go for runs? \\\hline
    \end{tabular}
    \begin{tabular}{c c}\\\hline
    \text{Persona 1} & \text{Persona 2} \\  \hline
    1. I am a psychology major. & 1. I don't pick up my toys.\\
    2. I enjoy broadway shows. & 2. I have a pet turtle.\\
    3. I'm a Steelers fan. & 3. I like to play with my dolls.\\
    4. My favorite band is the Avett Brother. & 4. My best friend lives next door.\\ \hline
    \end{tabular}
    \caption{An example of self-play between two randomized PersonaGPT models.}
    \label{tab:self-play}
\end{table}
\paragraph{Authenticator Policy} 
Table~\ref{tab:authenticator-personagpt} shows an example conversation between the learned authenticator model and a random PersonaGPT agent. Turn 1 and 8 highlight the importance of incorporating question codes as context for decoding: rather than hard-coding questions, an authenticator must also answer questions from the input agent (PersonaGPT) as to not disrupt the flow of conversation. 
Since the authenticator is not given any persona facts, it must make up an answer to PersonaGPT questions on the fly. 
For example, the authenticator states that its favorite color is green and that it is unemployed in order to answer queries in addition to proposing questions. 

In this particular conversation, the authenticator question codes are: \texttt{talk about work} for the first 3 turns, \texttt{ask about hobbies} for the next 2 turns, \texttt{talk about politics} for turn 6, \texttt{talk about travel} for turn 7, and \texttt{talk about age and gender} in turn 8. 
When question codes are repeated, the authenticator model is able to avoid the same question twice -- instead, it expands upon the existing knowledge in the conversation (e.g., ex-military, works in a shipyard). 
Note that when as context, the question codes function as one-to-many mappings. 
For example, ``talk about travel'' may correspond to a much larger set of topics other than traveling as a hobby. 
The authenciator may ask where the input agent lives, where they are from, etc. 
Similarly, ``ask about age and gender'' has much more to do with questions related to identity and appearance rather than strictly age and gender alone. 
\begin{table}[ht]
    \centering
    \small
    \begin{tabular}{ c c c }\\\hline 
        Role  &  Response \\ \hline
        Authenticator  &  hello there what kind of work do you do?\\
        PersonaGPT &  hi there i am ex military. what do you do for a living?\\\hline
        Authenticator & nothing much. what do you like about your job?\\
        PersonaGPT & i am a shipyard worker. \\\hline
        Authenticator  & what's your dream job?\\
        PersonaGPT &  to be a navy seal one day.\\\hline
        Authenticator  & that sounds exciting! do you have any hobbies?\\
        PersonaGPT &  my hobby is volunteering with the bunny rescue.\\\hline 
        Authenticator & do you have any other hobbies?\\
        PersonaGPT & yes i help out at the local bunny rescue.\\\hline
        Authenticator  & do you follow politics? \\
        PersonaGPT &  yes i follow politics. i donate a part of my salary to charity each month. \\\hline   
        Authenticator  & where are you from?\\
        PersonaGPT &  i'm from the united states. what's your favorite color?\\\hline 
        Authenticator  & my favorite color is green. how old are you?\\
        PersonaGPT & i'm 20 what about you? \\\hline

    \end{tabular}
    \begin{tabular}{c c}\\\hline
     \multicolumn{2}{c}{\text{Persona Facts}} \\  \hline
    1. I donate a part of my salary to charity each month. &
    2. I volunteer my time with a local bunny rescue.\\
    3. I work fulltime in a shipyard. &
    4. I'm ex military.\\ 
    5. My favorite color is navy blue.\\\hline
    \end{tabular}
    \caption{An example of interaction between the learned authenticator policy and PersonaGPT over an 8-turn conversation.}
    \label{tab:authenticator-personagpt}
\end{table}

\paragraph{Consent}
We obtained IRB approval for the data collection. When users register for the experiment, the following agreement and potential risk statements are required to be read and signed off on before user registration. 

\textit{You will be asked to engage in an 8-10 turn, short dialog with a chatbot. The chatbot has 4-5 personality traits, which will be revealed to you at the start of the experiment. At the end of the conversation, you will be asked to rate the chatbot on a series of performance metrics.}

\textit{Purpose of This Experiment. 
The purpose of this research study is to evaluate the performance of several state-of-the-art open-domain conversational agents developed through various learning algorithms. Automatic evaluation of conversational agents has proven to be difficult, and thus human evaluation is necessary to assess the performance of these agents.}

\textit{Your rights to participate, say no, or withdraw. 
Participation is voluntary, you may choose not to participate at all, or you may refuse to participate in certain procedures or answer certain questions or discontinue your participation at any time without consequence.}

\section{Model and Training Details}
All models were written using PyTorch~\cite{paszke2017automatic}. 
The PersonaGPT model was written using the HuggingFace Transformers package~\cite{wolf-etal-2020-transformers}. 
In terms of GPU usage, all models were trained using a single 11Gb NVIDIA GTX 1080 Ti. 
For experiment 4.1, the GPT-2 baseline, DialoGPT and PersonaGPT were fine-tuned on the \texttt{PersonaChat} dataset for 3 epochs, each taking between 13-16 hours of wall clock time. 
For each model, the AdamW~\cite{loshchilov2019decoupled} was used with an initial learn rate of 5e-5 and a linear decay schedule. 

For experiment 4.2, all identifier models were trained for 10 epochs. The BoW feed forward network (MLP) consisted of 2 hidden layers, 300 units each and a dropout rate of 0.2 between the layers. 
BoE, the MLP architecture consisted of 2 layers, 1024 units each with a dropout rate of 0.2 between layers. 
For the LSTM model, the input embedding size is 30, 1 LSTM layer is used with 600 hidden units. 
For the BERT and GPT-2 models, the transformer (feature representation) layers were frozen, and additional 2-layer MLP modules were added to each model for training, each consisting of 1024 units per layer. 
The identifier model is a 2-layer MLP with 1024 units each with a dropout rate of 0.2 between layers. 
All identifier models were trained using Adam~\cite{kingma2017adam} optimizer with learn rate of 1e-3.

The verifier network consists of a 3-layer MLP with 512 hidden units and dropout rate of 0.1 between layers. 
Tanh activation is used in place of ReLU, as we found Tanh to empirically outperform the latter in our use case. 
Note that the output layer size is 11 (corresponding to the size of the action space, i.e., number of question codes). 
This output layer is trained to fit the Q-targets during Q-learning, and an additional softmax layer is added to shape the Q-values into a probability distribution from which to sample the actions for decoding responses. 
The verifier network was pre-trained on the active learning data over 10,790 conversational turns for 3 epochs, totally between 3.5-4 wall clock hours. 
For Q-learning, the verifier was trained for 3 simulated conversations per training set persona, totalling 22 hours of wall clock time over 3,846 total conversations and 30,768 conversational turns. 
After each conversation during the DQN training loop, the Q-value layers are fine-tuned over the replay buffer for 3 epochs. 
For SWA, at the end of each gradient update for $\theta$, the target network is updated according to:
\begin{align}
    \theta' \leftarrow \eta \theta + (1-\eta) \theta',
    \label{eqn:swa}
\end{align}
where $\eta = 1/(N+1)$ and $N$ is the number of training iterations.
For $\varepsilon$-greedy, we set the initial $\varepsilon_0 = 0.5$, $\varepsilon_{\text{min}} = 0.05$, and decay factor to 2048. 

\subsection{Active Learning}
For the gradient descent step, we split the parameters of PersonaGPT ($\theta_{LM}$) into 4 groups: fast group (consisting of special tokens), slow group (consisting of positional codes), freeze group (embedding weights for normal tokens), and the rest of the parameters. 
We set the initial learn rates of each group as follows: fast group ($\alpha $ = 5e-4), slow group (1e-6), freeze group (1e-9), and the rest (5e-5). 
However, the full Fischer Information matrix is intractable to learn explicitly; instead, we design a diagonal matrix $M$, with entries corresponding to the learn rates of the different groups (4 different initial rates). 
The gradient descent update is then: 
$$\theta_{LM} \leftarrow \theta_{LM} - M \nabla_{\theta_{LM}} J(\theta_{LM})$$
Empirically, we find that this scheme allows PersonaGPT to incorporate question codes without sacrificing felicity of decoded responses. 

\subsection{Explanation of Prefix Codes} 
In total there are 11 possible discrete actions that the authenticator network can output. 
Each action corresponds to a particular phrase to be incorporated as prefix to PersonaGPT. 
However, PersonaGPT can take arbitruary persona information for conditional decoding. 
Consider the following toy example:\\
\noindent
{\small 
\texttt{<|p1|>I like dogs.<|sep|> <|start|> hi! how are you doing today?<|eos|>}}

The prefix code starts with \texttt{<|p1|>} and ends with \texttt{<|sep|>} to denote the persona input $P$. The text following \texttt{<|start|>} denotes the conditional decoding targets of the LM.
\section{Ablation Study}
In addition to the PersonaGPT model, we are also interested in the performance of the authenticator policy against other input models. 
For example, how well does our policy fair against models with less capacity to incorporate persona information? What about against models with lower decoding quality? 
We generate several synthetic conversations between our authenticator and several variations of persona models:
\begin{itemize}[leftmargin=0.1in]
    \item Full Persona: full persona model.
    \item Weak Persona: persona model with higher nucleus sampling size ($p \in [0.30-0.8]$)~\cite{holtzman2019curious} to capture less sensible models.
    \item Transition Model: model with either randomly initialized or no persona inputs (defaults to non-personalized decoding).
\end{itemize}
We use the transition model to serve as a baseline in which persona information is not incorporated in the input dialogue agent. 
Additionally, we include a ``weak persona'' model baseline, which in corporates persona information but suffers from decreased overall felicity. 
We randomly sample persona inputs from the full set of $1,412$ personas and report the mean prec@k and rec@k performance across generated conversations. 
%
Table~\ref{tab:ablation} compares authenticator performance against these persona model variants. 
As expected, the non-personalized transition model did not conform to given persona profiles, and the authenticator was most affected by the drop in personalization. 
By contrast, the authenticator was still able to maintain some performance against a much less felicitous persona model. 
\begin{table}[ht]
    \small
    \centering
    \begin{tabular}{c c c c c }\hline
         \textbf{Input Model} & \textbf{Prec@1} & \textbf{Prec@5} & \textbf{Rec@5} & \textbf{Rec@10} \\ \hline
         Transition & 17.1 & 17.1 & 19.4 & 40.6 \\
         Weak Persona & 79.8 & 49.5 & 55.9 & 74.1 \\ 
         Full Persona & \textbf{86.0} & \textbf{53.2} & \textbf{60.0} & \textbf{77.8} \\ \hline
    \end{tabular}
    \caption{Authenticator performance against variations of the input persona model.}
    \label{tab:ablation}
\end{table}

\section{Social Impact}
\label{sec:broad}
Beyond evaluating persona models, persona verification can be generalized to 
the setting of speaker verification for conversational agents, human or chatbot.  
In many real-world settings, speaker information such as audio and video may not be readily available. 
In such cases, the verifier network provides a way of speaker identification via text. 
One can think of persona verification as a way of obtaining a linguistic ``fingerprint'' of speakers based on the manner in which they converse under different question policies. 
For example, human speakers seeking access to personal data may go through a short conversation with the verifier network in order to see whether the person trying to access private data has the correct identity.

Verification is a critical issue in the modern era of cybersecurity. 
Consider for example the arrival of \emph{Deep Fakes}~\cite{guera2018deepfake} -- synthetically generated videos of people doing actions that may be outside the context of their persona. 
The use of only voice and video identifiers may not be enough to truly assess whether a person's physical features match with their actions. 
In this regard, we try to introduce the idea that the problem of verification may entail much more than just matching physical / biological features. 
This is why we approach the problem from the point of a \emph{question policy}, a verification process that is dynamic and stochastic rather than static and deterministic. 
In the latter case, technical advances in modern AI can ``game'' physical features which are fixed points in some classifier space. 
In the former case, however, an impersonator must do much more. 
 To fool a dynamic verifier such as ours, one has to find fixed points in \emph{policy space}, which involves sequential decision-making rather than one-time classification using facial recognition and fingerprint features. 

Beyond the positive impacts, there are numerous potential avenues for misuse of the proposed technology. We list some notable ones below:
\begin{itemize}
    \item Mistakes in persona identification can result in mistakes in granting / denying services for persons or groups of persons. For example, persona facts (or sets) for which the persona identifier possesses higher error rates can potentially lead to poor access for those potential users. 
    
    \item Similarly, verifier errors (e.g., poor questions delivered) with certain actions (e.g., talk about hobbies, talk about travel) may have disproportionate less consequences compared to more sensitive topics (e.g., talk about gender, talk about politics). 
    
    \item Although the verifier is meant to do authentication, it can potentially be abused to conduct conversations for the purposes of \emph{mining} persona information. 
    For example, an application using the verifier can abuse building rapport with human users to mine personal information. 
    We did not explore ways to prevent this type of misuse, but future work must focus on either counter-measures or methods of prevention against such cases.
    
    \item Algorithmic authentication and persona modeling can potentially greatly accelerate the development of human-like dialog generation.  Deployment considerations of conversational authentication must carefully consider the impact of persona modeling on the potential increase in the capacity for general chatbots to conduct deceptive / exploitative interactions (e.g., impersonation, personalized advertising, political manipulation) and their potentially detrimental impact on human labor conditions. 
    
\end{itemize}
Additionally, note that since the persona models used in our experiments are built from a language model pretrained on large-scale datasets, they have been shown to contain various cultural biases \cite{tamkin2021understanding,solaiman2019release}. Finetuning on PersonaChat certainly do not alleviate these issues, as the personas themselves were not curated against such biases. 
For example, the term ``gender'' used in this study is defined as gender \emph{perceived by the annotators of the PersonaChat dataset}. Its interpretation may not generalize to other real world settings. 

\bibliographystyle{unsrt}
